\documentclass[table]{hfstyle/hf}

\usepackage{microtype}
\usepackage{hyperref}
\usepackage{url}
\usepackage{booktabs}
\usepackage{graphicx}
\usepackage{lineno}
\usepackage{enumitem}
\usepackage{listings} %
\usepackage{svg}

\definecolor{ darkblue}{rgb}{0, 0, 0.5}
\hypersetup{colorlinks=true, citecolor=darkblue, linkcolor=darkblue, urlcolor=darkblue}

\usepackage{amssymb}
\usepackage{multirow}
\usepackage{bigdelim}
\usepackage{todonotes}
\usepackage{longtable}
\usepackage{tabularray}
\usepackage{wrapfig}
\usepackage[most]{tcolorbox} 
\usepackage{url}
\usepackage{xspace}
\usepackage{svg}
\usepackage[absolute]{textpos} %

\usepackage{fdsymbol}   %

\usepackage[utf8]{inputenc} %
\usepackage[T1]{fontenc}    %
\usepackage{url}            %
\usepackage{booktabs}       %
\usepackage{amsfonts}       %
\usepackage{nicefrac}       %
\usepackage{microtype}      %
\usepackage[table]{xcolor}         %
\usepackage{amsmath}
\usepackage[most]{tcolorbox}
\usepackage{csquotes}

\usepackage{siunitx}
\usepackage{graphicx}
\usepackage{arydshln}
\usepackage{wrapfig}
\usepackage{enumitem}
\usepackage{soul} %

\usepackage{multirow}
\usepackage{xspace}
\usepackage{adjustbox}
\usepackage{pifont}
\usepackage{caption}
\usepackage{makecell}
\usepackage{bold-extra}
\usepackage{url}

\usepackage{pgf-pie}

\usepackage{hyperref}
\definecolor{linkcolor}{RGB}{0, 0, 128}
\hypersetup{
     colorlinks   = true,
     citecolor    = linkcolor,
     linkcolor    = linkcolor,
     urlcolor     = linkcolor,
}
\usepackage{pifont}%
\usepackage{listings}
\setlist[itemize]{leftmargin=*,itemsep=0em,parsep=0.3em,topsep=0.3em}

\definecolor{maroon}{HTML}{F26035}
\definecolor{yellow}{HTML}{FDBC42}
\definecolor{lavender}{HTML}{734f96}
\definecolor{darkergrey}{HTML}{444444}
\definecolor{midgrey}{HTML}{e6eded}

\definecolor{neutralEight}{HTML}{343434}
\definecolor{neutralFive}{HTML}{838383}
\definecolor{neutralThree}{HTML}{bebebe}
\definecolor{neutralOne}{HTML}{dedede}
\definecolor{lightgrey}{HTML}{fafcfc}

\usepackage{tikz}

\definecolor{maroon}{HTML}{F26035}
\definecolor{yellow}{HTML}{FDBC42}
\definecolor{darkred}{RGB}{156, 39, 33}
\definecolor{darkblue}{RGB}{31, 90, 153}
\definecolor{forestgreen}{rgb}{0.13, 0.55, 0.13}
\definecolor{olmoDarkBlue}{HTML}{012e59}
\definecolor{olmoBlue}{HTML}{265ed4}
\definecolor{olmoLightBlue}{HTML}{012e59}
\definecolor{olmoTeal}{HTML}{00d5ff}
\definecolor{olmoYellow}{HTML}{ffbb00}
\definecolor{olmoOrange}{HTML}{ff9100}

\usepackage{setspace}

\usepackage{nicematrix}

\title{SAM2-UNeXT: An Improved High-Resolution Baseline for Adapting Foundation Models to Downstream Segmentation Tasks}

\authorOne[1*]{Xinyu Xiong}
\authorOne[2*]{Zihuang Wu}
\authorOne[1]{Lei Zhang}
\authorOne[3]{Lei Lu}
\authorOne[4]{Ming Li}
\authorOne[1**]{Guanbin Li}

\contribution[1]{Sun Yat-sen University}
\contribution[2]{Jiangxi Normal University}
\contribution[3]{Hainan University}
\contribution[4]{Shandong Inspur Database Technology Co., Ltd}
\contribution[]{*Equal Contribution}
\contribution[]{**Corresponding Author}

\abstract{Recent studies have highlighted the potential of adapting the Segment Anything Model (SAM) for various downstream tasks. However, constructing a more powerful and generalizable encoder to further enhance performance remains an open challenge. In this work, we propose SAM2-UNeXT, an advanced framework that builds upon the core principles of SAM2-UNet while extending the representational capacity of SAM2 through the integration of an auxiliary DINOv2 encoder. By incorporating a dual-resolution strategy and a dense glue layer, our approach enables more accurate segmentation with a simple architecture, relaxing the need for complex decoder designs. Extensive experiments conducted on four benchmarks, including dichotomous image segmentation, camouflaged object detection, marine animal segmentation, and remote sensing saliency detection, demonstrate the superior performance of our proposed method. The code is available at \url{https://github.com/WZH0120/SAM2-UNeXT}.}

\begin{document}
\maketitle

\section{Introduction}
Foundation models are playing an increasingly pivotal role in computer vision~\cite{vision_survey}, natural language processing~\cite{nlp_survey}, intelligent medicine~\cite{med_survey}, autonomous driving~\cite{drive_survey}, and other domains~\cite{MLLM_Survey,CVPR25_Head,LMM_Agent}. In the field of image segmentation, the Segment Anything Model (SAM)~\cite{ICCV23_SAM1,SAM2} family has sparked significant interest. Traditional small segmentation networks typically devote substantial design effort to complex decoder modules. However, a fundamental limitation persists: once the knowledge is lost in the encoding stage, it cannot be fully recovered during decoding. In contrast, foundation models leverage their large parameter capacity and sophisticated pretraining strategies to learn high-quality representations, enabling accurate segmentation performance even with relatively simple decoder architectures.

Although foundation models demonstrate strong generalization capabilities, task-specific adaptation, such as parameter-efficient fine-tuning (PEFT)~\cite{arXiv24_PEFTSurvey}, remains important for many downstream applications~\cite{gatenet,AIR24_BiRefNet}. Recent methods have achieved promising results by incorporating lightweight adapters~\cite{ICML19_Adapter, ICCVW23_SAMAdapter}, LoRA modules~\cite{ICLR22_LoRA}, or similar components into the encoder, often in conjunction with decoder refinement strategies~\cite{InfoFus25_SAMamba, SPL25_HFSSAM2}. Nevertheless, relying solely on SAM still results in limited generalization in some scenarios. For example, linear probing~\cite{SAMantic} of the SAM encoder on the ImageNet~\cite{imagenet} classification yields significantly lower accuracy compared to other large models such as CLIP~\cite{ICML21_CLIP} and DINOv2~\cite{dinov2}. A plausible explanation is that SAM's class-agnostic segmentation pretraining induces a representational bias that favoring fine-grained local details while capturing limited global semantic context.

Building upon the above analysis, we propose SAM2-UNeXT, a unified and extensible framework that synergistically integrates multiple foundation models, including SAM2~\cite{SAM2} and DINOv2~\cite{dinov2}, to harness their complementary strengths in detail perception and semantic representation. The proposed SAM2-UNeXT offers the following key benefits:

\begin{itemize}
    \item \textbf{Simplicity.} SAM2-UNeXT simplifies any additional attention design and focuses on a lightweight and efficient encoder fusion strategy.

    \item \textbf{Scalability.} With support for dynamic resolution adjustment and flexible auxiliary encoder configurations, SAM2-UNeXT can be readily adapted to a broad range of downstream tasks.

    \item \textbf{Effectiveness.} Extensive experiments on four public benchmarks demonstrate that SAM2-UNeXT consistently achieves strong segmentation performance across diverse scenarios under limited training epochs.
\end{itemize}

\begin{figure*}[t]
    \centering
    \includegraphics[width=1.0\linewidth]{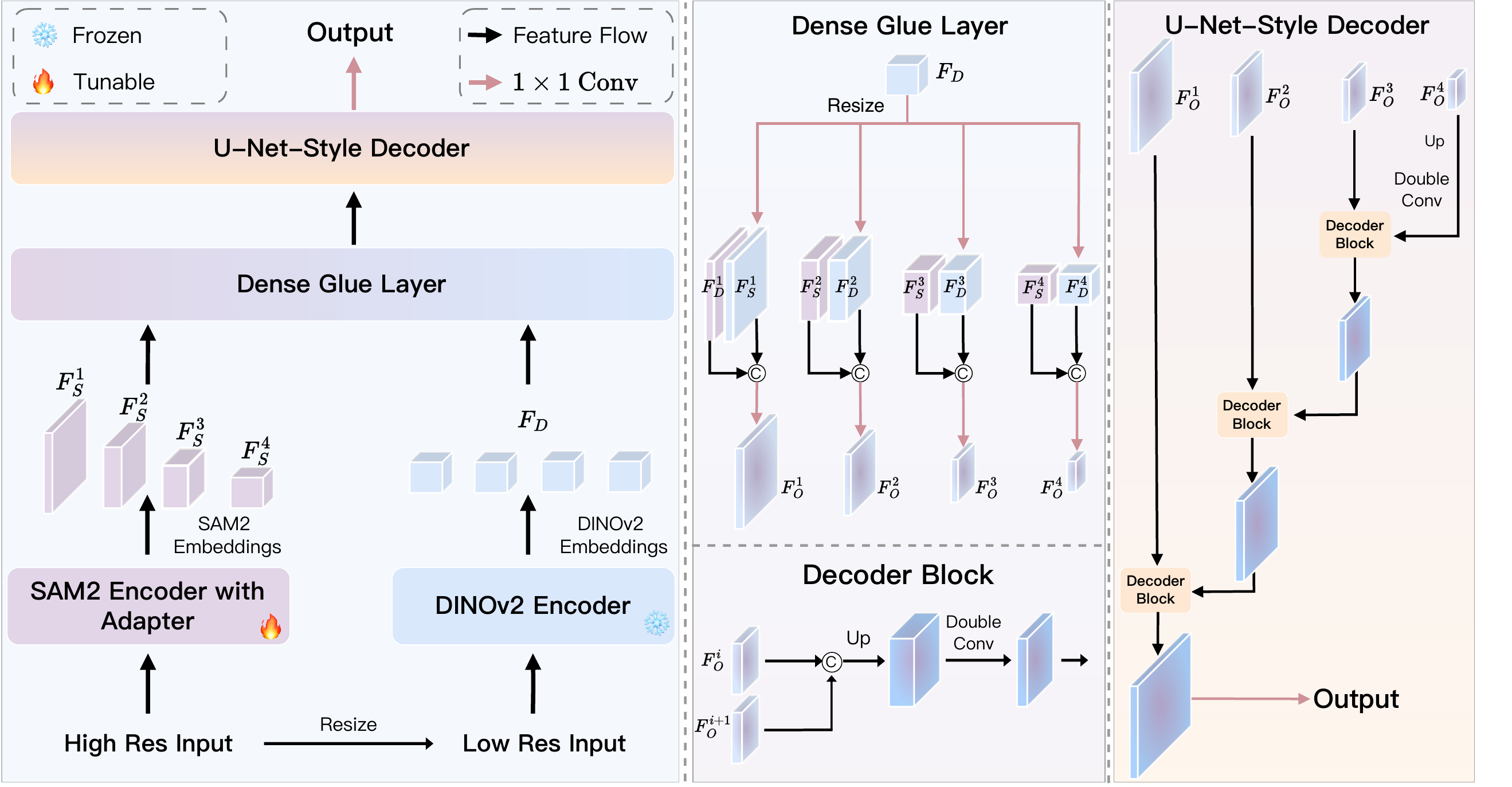}
    \caption{Overview of the proposed SAM2-UNeXT.} 
    \label{fig:sam2unext}
\end{figure*}

\section{Method}
As illustrated in Figure~\ref{fig:sam2unext}, the proposed architecture consists of four key components: a SAM2 encoder, a DINOv2 encoder, a dense glue layer, and a U-Net-style decoder.

\subsection{SAM2 Encoder}
In this stage, we closely follow the practice of SAM2-UNet~\cite{sam2unet}, adapting the Hiera~\cite{ICML23_Hiera} encoder from SAM2 and freezing all of its original parameters. Parameter-efficient fine-tuning (PEFT) is performed by inserting lightweight adapters~\cite{ICML19_Adapter} before each Hiera block. The adapter adopts a simple “MLP-GeLU-MLP-GeLU” structure with a 32-channel bottleneck.

\subsection{DINOv2 Encoder}
Compared with the Segment Anything series~\cite{ICCV23_SAM1,SAM2}, DINOv2~\cite{dinov2} serves as a more general-purpose vision foundation model. Trained through self-supervised learning, it demonstrates strong transferability across a wide range of vision tasks, including classification, segmentation, and depth estimation. In line with the original implementation, we freeze all DINOv2 parameters and do not apply any parameter-efficient fine-tuning strategies to balance training efficiency and performance.

\subsection{Dual-Resolution Design}
A straightforward approach to combining the two large encoders is to process inputs at the same resolution; however, this is computationally inefficient. In particular, for DINOv2, which relies on standard self-attention mechanisms~\cite{ICLR21_ViT}, increasing the input resolution leads to a substantial rise in computational cost. Considering that SAM focuses on fine-grained local details, whereas DINOv2 emphasizes global semantic understanding, we adopt a dual-resolution strategy: the SAM encoder operates on a higher-resolution input $(H_h, W_h)$, while the DINOv2 encoder processes a lower-resolution input $(H_l, W_l)$.

\subsection{Dense Glue Layer}
Unlike the hierarchical design of Hiera, the vanilla Vision Transformer~\cite{ICLR21_ViT} architecture adopted by DINOv2 produces non-hierarchical, scale-consistent embeddings at every layer. A common approach~\cite{arXiv25_DSUNet,MedIA24_TransUNet} to leverage such transformer features is to enhance the final feature map of a hierarchical encoder. Instead, we employ a dense fusion strategy inspired by the observation that DINOv2 exhibits strong zero-shot capabilities: its encoded representations become highly interpretable after principal component analysis, effectively highlighting the foreground of interest without any fine-tuning. In other words, these features can be regarded as spatial attention maps enriched with global semantic information.

Based on this, we first apply four 1×1 convolutions to align the channel dimension of the DINOv2 features (1024 channels for DINOv2-L) with those of the four stages of the SAM2 encoder features (144, 288, 576, and 1152 channels for Hiera-L). Next, the DINOv2 features are resized to match the spatial dimensions of each corresponding SAM2 feature map and fused via simple channel-wise concatenation. Finally, the concatenated features are compressed to 128 channels through 1×1 convolutions to improve training efficiency.

\subsection{U-Net-Style Decoder}
In this stage, we mostly follow the design of SAM2-UNet by replacing the original transformer-based decoder in SAM2 with a U-Net-style decoder~\cite{MICCAI15_UNet}, where each decoder block consists of two consecutive “Conv–BN–ReLU” layers. The major difference is that we introduce an additional partial decoder that without feature concatenation, resulting in a total of four decoder stages. This modification increases the resolution of the final segmentation feature map to one-half (rather than one-quarter) of the high-resolution input, which is advantageous for tasks that are sensitive to boundary segmentation accuracy.

\section{Experiment}
\subsection{Datasets and Benchmarks}
We conduct experiments on four public benchmarks spanning a range of segmentation tasks:

\textbf{Dichotomous Image Segmentation.}
We use the DIS5K~\cite{DIS5K} dataset for evaluation. The training set (DIS-TR) contains 3,000 images, while the evaluation is conducted on five subsets: DIS-VD (470), DIS-TE1 (500), DIS5K-TE2 (500), DIS-TE3 (500), and DIS-TE4 (500). Performance is measured using four metrics: S-measure ($S_\alpha$)~\cite{CVPR17_Smeasure}, weighted F-measure ($F_{\beta}^{w}$)~\cite{CVPR14_Fmeasure}, mean E-measure ($E_{\phi}$)~\cite{Emeasure}, and mean absolute error (MAE).

\textbf{Camouflaged Object Detection.}
We evaluate on four datasets: CHAMELEON~\cite{chameleon}, CAMO~\cite{CAMO}, COD10K~\cite{CVPR20_SINet}, and NC4K~\cite{CVPR21_NC4K}. The unified training set consists of 4,040 images (3,040 from COD10K and 1,000 from CAMO). The remaining CHAMELEON (76), CAMO (250), COD10K (2,026), and NC4K (4,121) images are used for testing. We report results using S-measure ($S_\alpha$), adaptive F-measure ($F_{\beta}$), mean E-measure ($E_{\phi}$), and mean absolute error (MAE).

\textbf{Marine Animal Segmentation.}
Two datasets are used for this task: MAS3K~\cite{TCSVT21_MAS3K}, with 1,769 training images and 1,141 test images; and RMAS~\cite{JOE23_MASNet}, with 2,514 training images and 500 test images. Evaluation is based on five metrics: mIoU, S-measure ($S_\alpha$), weighted F-measure ($F_{\beta}^{w}$), mean E-measure ($E_{\phi}$), and mean absolute error (MAE).

\textbf{Remote Sensing Saliency Detection.}
We use two datasets: EORSSD~\cite{TIP20_EORSSD}, with 1,400 training images and 600 test images; and ORSI-4199~\cite{ORSI-4199}, with 2,000 training images and 2,199 test images. Five metrics are used for evaluation: S-measure ($S_\alpha$), mean F-measure ($F_{\beta}^{mean}$), max F-measure ($F_{\beta}^{max}$), mean E-measure ($E_{\phi}$), and mean absolute error (MAE).

\begin{table*}[!htpb]
\centering
\caption{Results on dichotomous image segmentation.}
\label{tab:dis}
\renewcommand\arraystretch{1.2}
\renewcommand\tabcolsep{2pt}
\scalebox{1.0}{
\begin{tabular}{l|cccc|cccc|cccc}
\hline
    & 
    \multicolumn{4}{c|}{DIS-VD~\cite{DIS5K}} & \multicolumn{4}{c|}{DIS-TE1~\cite{DIS5K}} & 
    \multicolumn{4}{c}{DIS-TE2~\cite{DIS5K}} \\
    \multirow{-2}{*}{Methods} & $S_\alpha$ & $F_{\beta}^{w}$ & $E_{\phi}$ & MAE & $S_\alpha$ & $F_{\beta}^{w}$ & $E_{\phi}$ & MAE & $S_\alpha$ & $F_{\beta}^{w}$ & $E_{\phi}$ & MAE \\
    \hline
    U$^2$Net~\cite{PR20_U2Net} & 0.785 & 0.656 & 0.809 & 0.089 & 0.762 & 0.601 & 0.783 & 0.085 & 0.798 & 0.676 & 0.825 & 0.083\\
    HRNet~\cite{PAMI20_HRNet} & 0.767 & 0.641 & 0.824 & 0.095 & 0.742 & 0.579 & 0.797 & 0.088 & 0.784 & 0.664 & 0.840 & 0.087\\
    IS-Net~\cite{DIS5K} & 0.813 & 0.717 & 0.856 & 0.074 & 0.787 & 0.662 & 0.820 & 0.074 & 0.823 & 0.728 & 0.858 & 0.070\\
    UDUN~\cite{MM23_UDUN} & 0.838 & 0.763 & 0.892 & 0.059 & 0.817 & 0.720 & 0.864 & 0.059 & 0.843 & 0.768 & 0.886 & 0.058\\
    BiRefNet~\cite{AIR24_BiRefNet} & 0.898 & 0.854 & 0.931 & 0.038 & 0.885 & 0.819 & 0.911 & 0.037 & 0.900 & 0.857 & 0.930 & 0.036\\
    \hline
    \textbf{SAM2-UNeXT} & \textbf{0.910} & \textbf{0.864} & \textbf{0.938} & \textbf{0.034} & \textbf{0.892} & \textbf{0.829} & \textbf{0.917} & \textbf{0.034} & \textbf{0.916} & \textbf{0.873} & \textbf{0.941} & \textbf{0.030}\\ 
    
\hline \hline
    & 
    \multicolumn{4}{c|}{DIS-TE3~\cite{DIS5K}} & \multicolumn{4}{c|}{DIS-TE4~\cite{DIS5K}} & 
    \multicolumn{4}{c}{DIS-TE(1-4)~\cite{DIS5K}} \\
    \multirow{-2}{*}{Methods} & $S_\alpha$ & $F_{\beta}^{w}$ & $E_{\phi}$ & MAE & $S_\alpha$ & $F_{\beta}^{w}$ & $E_{\phi}$ & MAE & $S_\alpha$ & $F_{\beta}^{w}$ & $E_{\phi}$ & MAE \\
    \hline
    U$^2$Net~\cite{PR20_U2Net} & 0.823 & 0.721 & 0.856 & 0.073 & 0.814 & 0.707 & 0.837 & 0.085 & 0.799 & 0.676 & 0.825 & 0.082\\
    HRNet~\cite{PAMI20_HRNet} & 0.805 & 0.700 & 0.869 & 0.080 & 0.792 & 0.687 & 0.854 & 0.092 & 0.781 & 0.658 & 0.840 & 0.087\\
    IS-Net~\cite{DIS5K} & 0.836 & 0.758 & 0.883 & 0.064 & 0.830 & 0.753 & 0.870 & 0.072 & 0.819 & 0.726 & 0.858 & 0.070\\
    UDUN~\cite{MM23_UDUN} & 0.865 & 0.809 & 0.917 & 0.050 & 0.849 & 0.792 & 0.901 & 0.059 & 0.844 & 0.772 & 0.892 & 0.057\\
    BiRefNet~\cite{AIR24_BiRefNet} & 0.919 & 0.893 & 0.955 & 0.028 & 0.900 & 0.864 & 0.939 & 0.039 & 0.901 & 0.858 & 0.934 & 0.035\\ 
    \hline
    \textbf{SAM2-UNeXT} & \textbf{0.926} & \textbf{0.897} & \textbf{0.956} & \textbf{0.027} & \textbf{0.909} & \textbf{0.867} & \textbf{0.944} & \textbf{0.037} & \textbf{0.911} & \textbf{0.867} & \textbf{0.940} & \textbf{0.032}\\  
    \hline
\end{tabular}
}
\end{table*}

\begin{table*}[!htpb]
\centering
\caption{Results on camouflaged object detection.}
\label{tab:cod}
\renewcommand\arraystretch{1.2}
\renewcommand\tabcolsep{2pt}
\scalebox{0.85}{
\begin{tabular}{l|cccc|cccc|cccc|cccc}
\hline
    & 
    \multicolumn{4}{c|}{CHAMELEON~\cite{chameleon}} & \multicolumn{4}{c|}{CAMO~\cite{CAMO}} & 
    \multicolumn{4}{c|}{COD10K~\cite{CVPR20_SINet}} & \multicolumn{4}{c}{NC4K~\cite{CVPR21_NC4K}}\\
    \multirow{-2}{*}{Methods} & $S_\alpha$ & $F_{\beta}$ & $E_{\phi}$ & MAE & $S_\alpha$ & $F_{\beta}$ & $E_{\phi}$ & MAE & $S_\alpha$ & $F_{\beta}$ & $E_{\phi}$ & MAE & $S_\alpha$ & $F_{\beta}$ & $E_{\phi}$ & MAE\\
    \hline
    SINet~\cite{CVPR20_SINet} & 0.872 & 0.823 & 0.936 & 0.034 & 0.745 & 0.712 & 0.804 & 0.092 & 0.776 & 0.667 & 0.864 & 0.043 & 0.808 & 0.768 & 0.871 & 0.058 \\
    PFNet~\cite{CVPR21_PFNet} & 0.882 & 0.820 & 0.931 & 0.033 & 0.782 & 0.751 & 0.841 & 0.085 & 0.800 & 0.676 & 0.877 & 0.040 & 0.829 & 0.779 & 0.887 & 0.053 \\
    ZoomNet~\cite{CVPR22_ZoomNet} & 0.902 & 0.858 & 0.943 & 0.024 & 0.820 & 0.792 & 0.877 & 0.066 & 0.838 & 0.740 & 0.888 & 0.029 &  0.853 & 0.814 & 0.896 & 0.043\\
    FEDER~\cite{CVPR23_FEDER} & 0.903 & 0.856 & 0.947 & 0.026 & 0.836 & 0.807 & 0.897 & 0.066 & 0.844 & 0.748 &  0.911 & 0.029 & 0.862 &  0.824 &  0.913 &  0.042\\
    SAM2-UNet~\cite{sam2unet} & 0.914 & 0.863 & 0.961 & 0.022 & 0.884 & 0.861 & 0.932 & 0.042 & 0.880 & 0.789 & 0.936 & 0.021 & 0.901 & 0.863 & 0.941 & 0.029 \\  
    \hline
    \textbf{SAM2-UNeXT} & \textbf{0.942} & \textbf{0.916} & \textbf{0.972} & \textbf{0.013} & \textbf{0.903} & \textbf{0.891} & \textbf{0.941} & \textbf{0.036} & \textbf{0.924} & \textbf{0.867} & \textbf{0.964}  & \textbf{0.013} & \textbf{0.923}  & \textbf{0.903} & \textbf{0.954} & \textbf{0.022}\\  
    \hline
\end{tabular}
}
\end{table*}
\begin{table*}[!htpb]
\centering
\caption{Results on marine animal segmentation.}
\label{tab:mas}
\renewcommand\arraystretch{1.2}
\renewcommand\tabcolsep{2pt}
\begin{tabular}{l|ccccc|ccccc}
\hline
    & 
    \multicolumn{5}{c|}{MAS3K~\cite{TCSVT21_MAS3K}} & \multicolumn{5}{c}{RMAS~\cite{JOE23_MASNet}}
    \\
    \multirow{-2}{*}{Methods} & $mIoU$ & $S_\alpha$ & $F^{w}_{\beta}$ & $E_{\phi}$ & MAE & $mIoU$ & $S_\alpha$ & $F^{w}_{\beta}$ & $E_{\phi}$ & MAE\\
    \hline
    C2FNet~\cite{IJCAI21_C2FNet} & 0.717 & 0.851 & 0.761 & 0.894 & 0.038 & 0.721 & 0.858 & 0.788 & 0.923& 0.026 \\
    OCENet~\cite{WACV22_OCENet} & 0.667 & 0.824 & 0.703 & 0.868 & 0.052 & 0.680 & 0.836 & 0.752 & 0.900& 0.030\\
    ZoomNet~\cite{CVPR22_ZoomNet} & 0.736 & 0.862 & 0.780 & 0.898 & 0.032 & 0.728 & 0.855 & 0.795 & 0.915 & 0.022\\
    MASNet~\cite{JOE23_MASNet} & 0.742 & 0.864 & 0.788 & 0.906 & 0.032 & 0.731 & 0.862 & 0.801 & 0.920 & 0.024\\
    SAM2-UNet~\cite{sam2unet} & {0.799}	& {0.903} & {0.848} & {0.943} & {0.021} & {0.738} & {0.874} & {0.810} & {0.944} & {0.022}\\  
    \hline
    \textbf{SAM2-UNeXT} & \textbf{0.853}	& \textbf{0.926} & \textbf{0.900} & \textbf{0.960} & \textbf{0.014} & \textbf{0.774} & \textbf{0.883} & \textbf{0.841} & \textbf{0.949} & \textbf{0.019}\\  
    \hline
\end{tabular}
\end{table*}

\begin{table*}[!htpb]
\centering
\caption{Results on remote sensing saliency detection.}
\label{tab:ros}
\renewcommand\arraystretch{1.2}
\renewcommand\tabcolsep{2pt}
\begin{tabular}{l|ccccc|ccccc}
\hline
    & 
    \multicolumn{5}{c|}{ORSI-4199~\cite{ORSI-4199}} & \multicolumn{5}{c}{EORSSD~\cite{TIP20_EORSSD}} \\
    \multirow{-2}{*}{Methods} & $S_\alpha$ & $F_{\beta}^{mean}$ & $F_{\beta}^{max}$ & $E_{\phi}$ & MAE & $S_\alpha$ & $F_{\beta}^{mean}$ & $F_{\beta}^{max}$ & $E_{\phi}$ & MAE \\
    \hline
    EMFINet~\cite{TGRS22_EMFINet} & 0.859 & 0.810 & 0.817 & 0.902 & 0.045 & 0.932 & 0.851 & 0.874 & 0.960 & 0.008\\
    ACCoNet~\cite{TCYB22_ACCoNet} & 0.868 & 0.861 & 0.865 & 0.934 & 0.031 & 0.929 & 0.855 & 0.884 & 0.965 & 0.007  \\
    ERPNet~\cite{TCYB22_ERPNet} & 0.865 & 0.839 & 0.854 & 0.917 & 0.037 & 0.925 & 0.827 & 0.874 & 0.937 & 0.008 \\
    AESINet~\cite{TGRS23_AESINet} & 0.870 & 0.863 & 0.868 & 0.936 & 0.031 & 0.935 & 0.850 & 0.879 & 0.965 & 0.006 \\
    SFANet~\cite{TGRS24_SFANet} & 0.876 & 0.866 & 0.871 & 0.939 & 0.029 & 0.935 & 0.868 & 0.883 & 0.973 & 0.006\\ 
    \hline
    \textbf{SAM2-UNeXT} & \textbf{0.887} & \textbf{0.873} & \textbf{0.886} & \textbf{0.942} & \textbf{0.026} & \textbf{0.948} & \textbf{0.892} & \textbf{0.905} & \textbf{0.978} & \textbf{0.004}\\  
    \hline
\end{tabular}
\end{table*}

\subsection{Implementation Details}
Our method is implemented in PyTorch and trained on a NVIDIA RTX 4090 GPU with 24 GB of memory. We use the AdamW optimizer with an initial learning rate of 0.0002 and apply cosine learning rate decay to stabilize training. The overall loss function~\cite{AAAI20_F3Net} consists of a weighted cross-entropy loss ($\mathcal{L}^{\omega}_{\text{BCE}}$) and a weighted IoU loss ($\mathcal{L}^{\omega}_{\text{IoU}}$). Two data augmentation strategies, including random horizontal and vertical flipping, are employed during training. Unless otherwise specified, we adopt the large version of SAM2 and DINOv2. The input resolutions are set to $(H_h, W_h) = (1024, 1024)$ for the SAM2 branch and $(H_l, W_l) = (448, 448)$ for the DINOv2 branch. All models are trained with a batch size of 1 for 20 epochs across all tasks.

\begin{figure*}[t]
    \centering
    \includegraphics[width=1.0\linewidth]{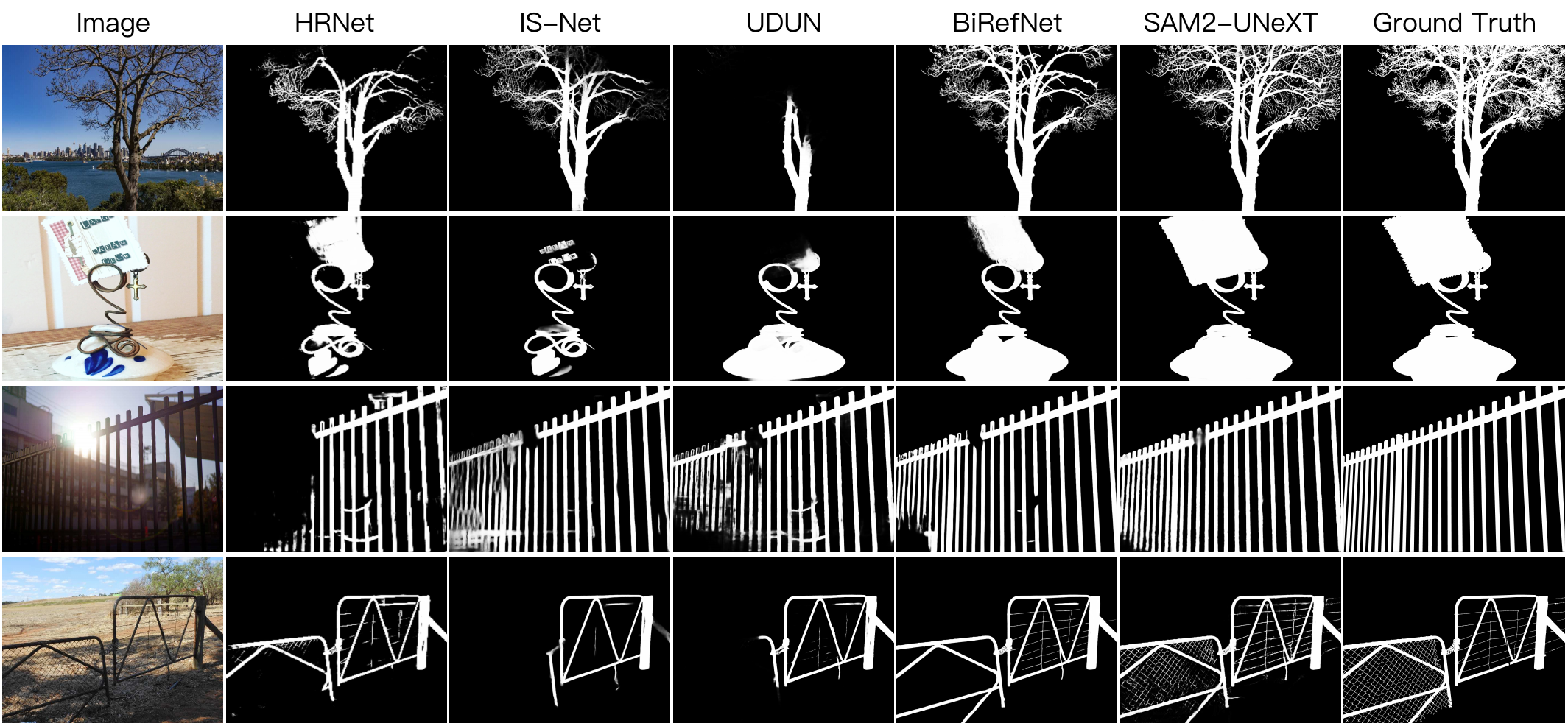}
    \caption{Visualization results on dichotomous image segmentation.} 
    \label{fig:dis5k}
\end{figure*}

\subsection{Comparison with State-of-the-Art Methods}
In this subsection, we first analyze the quantitative results across multiple benchmarks, followed by qualitative visual comparisons for dichotomous image segmentation.

\textbf{Dichotomous Image Segmentation.}
The results are presented in Table~\ref{tab:dis}, SAM2-UNeXT achieves steady performance gains over the second-best method, BiRefNet. Specifically, on the DIS-VD subset, our method improves the S-measure by 1.2\%.

\textbf{Camouflaged Object Detection.}
The results are presented in Table~\ref{tab:cod}. Compared with SAM2-UNet, the new SAM2-UNeXT achieves consistent improvements across all metrics. For example, on the CHAMELEON dataset, SAM2-UNeXT improves the S-measure by 2.8\%.

\textbf{Marine Animal Segmentation.}
The results are presented in Table~\ref{tab:mas}. SAM2-UNeXT significantly outperforms existing methods by a large margin. For instance, on the MAS3K dataset, our method improves the mIoU by 5.4\%.

\textbf{Remote Sensing Saliency Detection.}
The results are presented in Table~\ref{tab:ros}. SAM2-UNeXT outperforms all competing methods on both datasets. Notably, on the ORSI-4199 dataset, our method achieves a 1.1\% improvement in S-measure.

\textbf{Qualitative Comparison.}
Figure~\ref{fig:dis5k} illustrates visual comparisons on the dichotomous image segmentation task. Our method demonstrates superior segmentation accuracy across diverse scenarios: fine-grained tree branches (row 1), complex multi-object compositions (row 2), light variations (row 3), and scenes with grid structures and shadow interference (row 4). SAM2-UNeXT effectively handles curved edges, thin structures, and subtle visual boundaries, delivering better segmentation results even under challenging conditions.

\begin{table*}[t]
\centering
\begin{minipage}{0.48\textwidth}
\centering
\caption{Results on different auxiliary encoders.}
\label{tab:aux_encoder}
\renewcommand\arraystretch{1.2}
\renewcommand\tabcolsep{2pt}
\begin{tabular}{l|ccccc}
\hline
    & 
    \multicolumn{5}{c}{MAS3K~\cite{TCSVT21_MAS3K}}
    \\
    \multirow{-2}{*}{Methods} & $mIoU$ & $S_\alpha$ & $F^{w}_{\beta}$ & $E_{\phi}$ & MAE\\
    \hline
    w/o aux & 0.832 & 0.918 & 0.882 & 0.949 & 0.017\\
    ResNet-101~\cite{resnet} & 0.840 & 0.917 & 0.885 & 0.955 & 0.017 \\
    PVTv2-b5~\cite{pvtv2} & 0.833 & 0.917 & 0.878 & 0.952 & 0.018\\
    DINOv2-S~\cite{dinov2} & 0.836 & 0.917 & 0.881 & 0.950 & 0.017\\
    DINOv2-B~\cite{dinov2} & 0.843 & 0.921 & 0.890 & 0.955 & 0.015\\  
    \hline
    \textbf{DINOv2-L} & \textbf{0.853}	& \textbf{0.926} & \textbf{0.900} & \textbf{0.960} & \textbf{0.014} \\  
    \hline
\end{tabular}
\end{minipage}
\hfill
\begin{minipage}{0.48\textwidth}
\centering
\caption{Results on different resolutions.}
\label{tab:aux_resolution}
\renewcommand\arraystretch{1.2}
\renewcommand\tabcolsep{2pt}
\begin{tabular}{cc|ccccc} 
\hline
\multicolumn{2}{c|}{Resolution} & \multicolumn{5}{c}{MAS3K~\cite{TCSVT21_MAS3K}} \\
H & L & $mIoU$         & \multicolumn{1}{c}{$S_\alpha$}     & \multicolumn{1}{c}{$F^w_\beta$}    & \multicolumn{1}{c}{$E_\phi$}       & \multicolumn{1}{c}{MAE}             \\ 
\hline
352 & 352 &0.820 & 0.913 & 0.871 & 0.952 & 0.016\\ 
1024 & 224 & 0.842 & 0.920 &  0.888 & 0.957 & 0.015\\ 
1024 & 672 & \textbf{0.853} & 0.924 & 0.897 & \textbf{0.960} & \textbf{0.014}\\ 
\hline
\textbf{1024} & \textbf{448} & \textbf{0.853} & \textbf{0.926} & \textbf{0.900} & \textbf{0.960} & \textbf{0.014} \\
\hline
\end{tabular}
\end{minipage}
\end{table*}

\subsection{Discussion}
In this section, we analyze the design choices of SAM2-UNeXT using MAS3K as a representative benchmark.

\subsubsection{Impact of Auxiliary Encoder}
We investigate the effect of different auxiliary encoder designs, as shown in Table~\ref{tab:aux_encoder}:

\textbf{Row 1.} The auxiliary encoder is removed. In this setting, the model roughly becomes a high-resolution variant of SAM2-UNet~\cite{sam2unet}. Although it performs better than the low-resolution version of SAM2-UNet, its accuracy remains lower than that of configurations with an auxiliary encoder.

\textbf{Row 2 \& 3.} We use ResNet-101~\cite{resnet} and PVTv2-b5~\cite{pvtv2} as auxiliary encoders with parameters trainable. The results show marginal improvements compared to the setting without an auxiliary encoder, suggesting limited benefits from these conventional backbones under a simple fusion strategy.

\textbf{Row 4 \& 5.} We replace the auxiliary encoder with frozen small and base versions of DINOv2~\cite{dinov2}. The results indicate that larger variants generally yield better performance.

\subsubsection{Impact of Dynamic Resolution}
We also explore the effect of different resolution combinations, as shown in Table~\ref{tab:aux_resolution}:

\textbf{Row 1.} Both SAM2 and DINOv2 encoders operate at a uniform low resolution of 352×352. This setting results in the lowest performance among all tested configurations, though it still surpasses the original SAM2-UNet baseline.

\textbf{Row 2.} The high resolution is fixed at 1024×1024 for the SAM2 branch, while the low resolution for the DINOv2 branch is reduced to 224×224. A slight performance drop is observed compared to the 448×448 setting, but it still outperforms the uniform 352×352 case.

\textbf{Row 3.} The high resolution remains at 1024×1024, while the low resolution is increased to 672×672. The performance difference is negligible compared to the 448×448 setting, but inference cost increases significantly, making this configuration less practical.

\section{Related Work}
\subsection{Fusing Foundation Models}
Integrating different foundation models has become a common strategy in recent years. Many Vision-Language Models (VLMs)~\cite{internvl3,qwen2.5vl} are composed of a vision encoder paired with a Large Language Model (LLM), enabling flexible combinations tailored to diverse applications. For the SAM series, there have been several efforts~\cite{CVPRW24_SAMCLIP,CVPR24_AlignSAM,MICCAI24_TPDRSeg} to enhance language understanding by incorporating CLIP~\cite{ICML21_CLIP}. Other works have focused on improving few-shot segmentation capabilities by integrating pretrained vision encoders like DINOv2~\cite{dinov2}, exemplified by Matcher~\cite{matcher}. The study most related to ours is~\cite{arXiv25_DSUNet}, which also introduces an auxiliary DINOv2 encoder to form a U-shaped architecture. However, their focus lies in the design of more sophisticated decoder design, such as content-guided attention~\cite{context_atten} and wavelet convolution~\cite{wave_conv}.

\subsection{Image Segmentation}
Image segmentation, viewed as a pixel-level classification task, can be broadly categorized into binary segmentation~\cite{gatenet}, semantic segmentation~\cite{semantic_seg}, instance segmentation~\cite{instance_seg}, and panoptic segmentation~\cite{pan_seg}. This work focuses on binary segmentation, where all foreground pixels are assigned to a single class, and the remaining pixels are treated as background. Binary segmentation underpins many important application domains, including dichotomous image segmentation~\cite{AIR24_BiRefNet,MM23_UDUN}, camouflaged object detection~\cite{CVPR23_FEDER,SPL25_HFSSAM2}, marine animal segmentation~\cite{TCSVT21_MAS3K,JOE23_MASNet}, and remote sensing saliency detection~\cite{TGRS24_SFANet,TGRS23_AESINet}. Most existing methods tend to design task-specific decoders for each segmentation scenario. In contrast, our proposed method introduces a unified framework capable of achieving state-of-the-art performance across multiple binary segmentation tasks with a single model architecture.

\section{Conclusion}
In this paper, we presented SAM2-UNeXT, a simple yet effective framework that integrates two powerful foundation models: SAM2 and DINOv2, through a decoupled resolution strategy. This design leverages the complementary feature biases of each model, resulting in enhanced segmentation performance. Extensive experiments on four benchmark datasets demonstrate the effectiveness and generalizability of our approach. Moreover, SAM2-UNeXT offers high customizability, making it well-suited for adaptation to a wide range of downstream tasks. By adjusting the dynamic resolution configuration or incorporating alternative auxiliary encoders, the framework holds promise for extending SAM2-based models to previously underexplored segmentation scenarios.

\bibliographystyle{hfstyle/plainnat}
\bibliography{references}

\begin{thebibliography}{66}
\providecommand{\natexlab}[1]{#1}
\providecommand{\url}[1]{\texttt{#1}}
\expandafter\ifx\csname urlstyle\endcsname\relax
  \providecommand{\doi}[1]{doi: #1}\else
  \providecommand{\doi}{doi: \begingroup \urlstyle{rm}\Url}\fi

\bibitem[Awais et~al.(2025)Awais, Naseer, Khan, Anwer, Cholakkal, Shah, Yang, and Khan]{vision_survey}
Muhammad Awais, Muzammal Naseer, Salman Khan, Rao~Muhammad Anwer, Hisham Cholakkal, Mubarak Shah, Ming-Hsuan Yang, and Fahad~Shahbaz Khan.
\newblock Foundation models defining a new era in vision: a survey and outlook.
\newblock \emph{IEEE Transactions on Pattern Analysis and Machine Intelligence}, 2025.

\bibitem[Bai et~al.(2025)Bai, Chen, Liu, Wang, Ge, Song, Dang, Wang, Wang, Tang, et~al.]{qwen2.5vl}
Shuai Bai, Keqin Chen, Xuejing Liu, Jialin Wang, Wenbin Ge, Sibo Song, Kai Dang, Peng Wang, Shijie Wang, Jun Tang, et~al.
\newblock Qwen2. 5-vl technical report.
\newblock \emph{arXiv preprint arXiv:2502.13923}, 2025.

\bibitem[Chen et~al.(2024{\natexlab{a}})Chen, Mei, Li, Lu, Yu, Wei, Luo, Xie, Adeli, Wang, et~al.]{MedIA24_TransUNet}
Jieneng Chen, Jieru Mei, Xianhang Li, Yongyi Lu, Qihang Yu, Qingyue Wei, Xiangde Luo, Yutong Xie, Ehsan Adeli, Yan Wang, et~al.
\newblock Transunet: Rethinking the u-net architecture design for medical image segmentation through the lens of transformers.
\newblock \emph{Medical Image Analysis}, page 103280, 2024{\natexlab{a}}.

\bibitem[Chen et~al.(2023)Chen, Zhu, Deng, Cao, Wang, Zhang, Li, Sun, Zang, and Mao]{ICCVW23_SAMAdapter}
Tianrun Chen, Lanyun Zhu, Chaotao Deng, Runlong Cao, Yan Wang, Shangzhan Zhang, Zejian Li, Lingyun Sun, Ying Zang, and Papa Mao.
\newblock Sam-adapter: Adapting segment anything in underperformed scenes.
\newblock In \emph{ICCVW}, pages 3367--3375, 2023.

\bibitem[Chen et~al.(2024{\natexlab{b}})Chen, He, and Lu]{context_atten}
Zixuan Chen, Zewei He, and Zhe-Ming Lu.
\newblock Dea-net: Single image dehazing based on detail-enhanced convolution and content-guided attention.
\newblock \emph{IEEE Transactions on Image Processing}, 33:\penalty0 1002--1015, 2024{\natexlab{b}}.

\bibitem[Deng et~al.(2009)Deng, Dong, Socher, Li, Li, and Fei-Fei]{imagenet}
Jia Deng, Wei Dong, Richard Socher, Li-Jia Li, Kai Li, and Li~Fei-Fei.
\newblock Imagenet: A large-scale hierarchical image database.
\newblock In \emph{CVPR}, pages 248--255. IEEE, 2009.

\bibitem[Dosovitskiy et~al.(2021)Dosovitskiy, Beyer, Kolesnikov, Weissenborn, Zhai, Unterthiner, Dehghani, Minderer, Heigold, Gelly, Uszkoreit, and Houlsby]{ICLR21_ViT}
Alexey Dosovitskiy, Lucas Beyer, Alexander Kolesnikov, Dirk Weissenborn, Xiaohua Zhai, Thomas Unterthiner, Mostafa Dehghani, Matthias Minderer, Georg Heigold, Sylvain Gelly, Jakob Uszkoreit, and Neil Houlsby.
\newblock An image is worth 16x16 words: Transformers for image recognition at scale.
\newblock In \emph{ICLR}, 2021.

\bibitem[Espinosa et~al.(2024)Espinosa, Yang, Ericsson, McDonagh, and Crowley]{SAMantic}
Miguel Espinosa, Chenhongyi Yang, Linus Ericsson, Steven McDonagh, and Elliot~J Crowley.
\newblock There is no samantics! exploring sam as a backbone for visual understanding tasks.
\newblock \emph{arXiv preprint arXiv:2411.15288}, 2024.

\bibitem[Fan et~al.(2017)Fan, Cheng, Liu, Li, and Borji]{CVPR17_Smeasure}
Deng-Ping Fan, Ming-Ming Cheng, Yun Liu, Tao Li, and Ali Borji.
\newblock Structure-measure: A new way to evaluate foreground maps.
\newblock In \emph{ICCV}, pages 4548--4557, 2017.

\bibitem[Fan et~al.(2020)Fan, Ji, Sun, Cheng, Shen, and Shao]{CVPR20_SINet}
Deng-Ping Fan, Ge-Peng Ji, Guolei Sun, Ming-Ming Cheng, Jianbing Shen, and Ling Shao.
\newblock Camouflaged object detection.
\newblock In \emph{CVPR}, pages 2777--2787, 2020.

\bibitem[Fan et~al.(2021)Fan, Ji, Qin, and Cheng]{Emeasure}
Deng-Ping Fan, Ge-Peng Ji, Xuebin Qin, and Ming-Ming Cheng.
\newblock Cognitive vision inspired object segmentation metric and loss function.
\newblock \emph{Scientia Sinica Informationis}, 6\penalty0 (6):\penalty0 5, 2021.

\bibitem[Finder et~al.(2024)Finder, Amoyal, Treister, and Freifeld]{wave_conv}
Shahaf~E Finder, Roy Amoyal, Eran Treister, and Oren Freifeld.
\newblock Wavelet convolutions for large receptive fields.
\newblock In \emph{ECCV}, pages 363--380. Springer, 2024.

\bibitem[Fu et~al.(2023)Fu, Chen, Huang, Cheng, Ding, and Ma]{JOE23_MASNet}
Zhenqi Fu, Ruizhe Chen, Yue Huang, En~Cheng, Xinghao Ding, and Kai-Kuang Ma.
\newblock Masnet: A robust deep marine animal segmentation network.
\newblock \emph{IEEE Journal of Oceanic Engineering}, 2023.

\bibitem[Gu et~al.(2022)Gu, Bai, and Kong]{instance_seg}
Wenchao Gu, Shuang Bai, and Lingxing Kong.
\newblock A review on 2d instance segmentation based on deep neural networks.
\newblock \emph{Image and Vision Computing}, 120:\penalty0 104401, 2022.

\bibitem[Hao et~al.(2020)Hao, Zhou, and Guo]{semantic_seg}
Shijie Hao, Yuan Zhou, and Yanrong Guo.
\newblock A brief survey on semantic segmentation with deep learning.
\newblock \emph{Neurocomputing}, 406:\penalty0 302--321, 2020.

\bibitem[He et~al.(2023)He, Li, Zhang, Tang, Zhang, Guo, and Li]{CVPR23_FEDER}
Chunming He, Kai Li, Yachao Zhang, Longxiang Tang, Yulun Zhang, Zhenhua Guo, and Xiu Li.
\newblock Camouflaged object detection with feature decomposition and edge reconstruction.
\newblock In \emph{CVPR}, pages 22046--22055, 2023.

\bibitem[He et~al.(2016)He, Zhang, Ren, and Sun]{resnet}
Kaiming He, Xiangyu Zhang, Shaoqing Ren, and Jian Sun.
\newblock Deep residual learning for image recognition.
\newblock In \emph{CVPR}, pages 770--778, 2016.

\bibitem[Houlsby et~al.(2019)Houlsby, Giurgiu, Jastrzebski, Morrone, De~Laroussilhe, Gesmundo, Attariyan, and Gelly]{ICML19_Adapter}
Neil Houlsby, Andrei Giurgiu, Stanislaw Jastrzebski, Bruna Morrone, Quentin De~Laroussilhe, Andrea Gesmundo, Mona Attariyan, and Sylvain Gelly.
\newblock Parameter-efficient transfer learning for nlp.
\newblock In \emph{ICML}, pages 2790--2799. PMLR, 2019.

\bibitem[Hu et~al.(2022)Hu, Shen, Wallis, Allen-Zhu, Li, Wang, Wang, Chen, et~al.]{ICLR22_LoRA}
Edward~J Hu, Yelong Shen, Phillip Wallis, Zeyuan Allen-Zhu, Yuanzhi Li, Shean Wang, Lu~Wang, Weizhu Chen, et~al.
\newblock Lora: Low-rank adaptation of large language models.
\newblock \emph{ICLR}, 2022.

\bibitem[Huang et~al.(2024)Huang, Xiong, Ma, Li, Jie, Ma, and Li]{CVPR24_AlignSAM}
Duojun Huang, Xinyu Xiong, Jie Ma, Jichang Li, Zequn Jie, Lin Ma, and Guanbin Li.
\newblock Alignsam: Aligning segment anything model to open context via reinforcement learning.
\newblock In \emph{CVPR}, pages 3205--3215, 2024.

\bibitem[Khan et~al.(2025)Khan, Leem, See, Wong, Zhang, and Fang]{med_survey}
Wasif Khan, Seowung Leem, Kyle~B See, Joshua~K Wong, Shaoting Zhang, and Ruogu Fang.
\newblock A comprehensive survey of foundation models in medicine.
\newblock \emph{IEEE Reviews in Biomedical Engineering}, 2025.

\bibitem[Kirillov et~al.(2023)Kirillov, Mintun, Ravi, Mao, Rolland, Gustafson, Xiao, Whitehead, Berg, Lo, et~al.]{ICCV23_SAM1}
Alexander Kirillov, Eric Mintun, Nikhila Ravi, Hanzi Mao, Chloe Rolland, Laura Gustafson, Tete Xiao, Spencer Whitehead, Alexander~C Berg, Wan-Yen Lo, et~al.
\newblock Segment anything.
\newblock In \emph{ICCV}, pages 4015--4026, 2023.

\bibitem[Lai et~al.(2025)Lai, Zhong, Qin, Ren, Wang, and Li]{CVPR25_Head}
Peiwen Lai, Weizhi Zhong, Yipeng Qin, Xiaohang Ren, Baoyuan Wang, and Guanbin Li.
\newblock Llm-driven multimodal and multi-identity listening head generation.
\newblock In \emph{CVPR}, pages 10656--10666, 2025.

\bibitem[Le et~al.(2019)Le, Nguyen, Nie, Tran, and Sugimoto]{CAMO}
Trung-Nghia Le, Tam~V Nguyen, Zhongliang Nie, Minh-Triet Tran, and Akihiro Sugimoto.
\newblock Anabranch network for camouflaged object segmentation.
\newblock \emph{Computer Vision and Image Understanding}, 184:\penalty0 45--56, 2019.

\bibitem[Li et~al.(2022)Li, Liu, Zeng, Lin, and Ling]{TCYB22_ACCoNet}
Gongyang Li, Zhi Liu, Dan Zeng, Weisi Lin, and Haibin Ling.
\newblock Adjacent context coordination network for salient object detection in optical remote sensing images.
\newblock \emph{IEEE Transactions on Cybernetics}, 53\penalty0 (1):\penalty0 526--538, 2022.

\bibitem[Li et~al.(2021)Li, Dong, Rigall, Zhou, Dong, and Chen]{TCSVT21_MAS3K}
Lin Li, Bo~Dong, Eric Rigall, Tao Zhou, Junyu Dong, and Geng Chen.
\newblock Marine animal segmentation.
\newblock \emph{IEEE Transactions on Circuits and Systems for Video Technology}, 32\penalty0 (4):\penalty0 2303--2314, 2021.

\bibitem[Li et~al.(2024)Li, Xiong, Xia, Ju, and Ge]{MICCAI24_TPDRSeg}
Wenxue Li, Xinyu Xiong, Peng Xia, Lie Ju, and Zongyuan Ge.
\newblock Tp-drseg: Improving diabetic retinopathy lesion segmentation with explicit text-prompts assisted sam.
\newblock In \emph{MICCAI}, pages 743--753. Springer, 2024.

\bibitem[Li and Chen(2022)]{pan_seg}
Xinye Li and Ding Chen.
\newblock A survey on deep learning-based panoptic segmentation.
\newblock \emph{Digital Signal Processing}, 120:\penalty0 103283, 2022.

\bibitem[Liu et~al.(2022)Liu, Zhang, and Barnes]{WACV22_OCENet}
Jiawei Liu, Jing Zhang, and Nick Barnes.
\newblock Modeling aleatoric uncertainty for camouflaged object detection.
\newblock In \emph{WACV}, pages 1445--1454, 2022.

\bibitem[Liu et~al.(2024)Liu, Zhu, Li, Chen, Wang, and Shen]{matcher}
Yang Liu, Muzhi Zhu, Hengtao Li, Hao Chen, Xinlong Wang, and Chunhua Shen.
\newblock Matcher: Segment anything with one shot using all-purpose feature matching.
\newblock In \emph{ICLR}, 2024.

\bibitem[Lv et~al.(2021)Lv, Zhang, Dai, Li, Liu, Barnes, and Fan]{CVPR21_NC4K}
Yunqiu Lv, Jing Zhang, Yuchao Dai, Aixuan Li, Bowen Liu, Nick Barnes, and Deng-Ping Fan.
\newblock Simultaneously localize, segment and rank the camouflaged objects.
\newblock In \emph{CVPR}, pages 11591--11601, 2021.

\bibitem[Margolin et~al.(2014)Margolin, Zelnik-Manor, and Tal]{CVPR14_Fmeasure}
Ran Margolin, Lihi Zelnik-Manor, and Ayellet Tal.
\newblock How to evaluate foreground maps?
\newblock In \emph{CVPR}, pages 248--255, 2014.

\bibitem[Mei et~al.(2021)Mei, Ji, Wei, Yang, Wei, and Fan]{CVPR21_PFNet}
Haiyang Mei, Ge-Peng Ji, Ziqi Wei, Xin Yang, Xiaopeng Wei, and Deng-Ping Fan.
\newblock Camouflaged object segmentation with distraction mining.
\newblock In \emph{CVPR}, pages 8772--8781, 2021.

\bibitem[Naveed et~al.(2023)Naveed, Khan, Qiu, Saqib, Anwar, Usman, Akhtar, Barnes, and Mian]{nlp_survey}
Humza Naveed, Asad~Ullah Khan, Shi Qiu, Muhammad Saqib, Saeed Anwar, Muhammad Usman, Naveed Akhtar, Nick Barnes, and Ajmal Mian.
\newblock A comprehensive overview of large language models.
\newblock \emph{ACM Transactions on Intelligent Systems and Technology}, 2023.

\bibitem[Oquab et~al.(2024)Oquab, Darcet, Moutakanni, Vo, Szafraniec, Khalidov, Fernandez, HAZIZA, Massa, El-Nouby, et~al.]{dinov2}
Maxime Oquab, Timoth{\'e}e Darcet, Th{\'e}o Moutakanni, Huy~V Vo, Marc Szafraniec, Vasil Khalidov, Pierre Fernandez, Daniel HAZIZA, Francisco Massa, Alaaeldin El-Nouby, et~al.
\newblock Dinov2: Learning robust visual features without supervision.
\newblock \emph{Transactions on Machine Learning Research}, 2024.

\bibitem[Pang et~al.(2022)Pang, Zhao, Xiang, Zhang, and Lu]{CVPR22_ZoomNet}
Youwei Pang, Xiaoqi Zhao, Tian-Zhu Xiang, Lihe Zhang, and Huchuan Lu.
\newblock Zoom in and out: A mixed-scale triplet network for camouflaged object detection.
\newblock In \emph{CVPR}, pages 2160--2170, 2022.

\bibitem[Pei et~al.(2023)Pei, Zhou, Jin, Tang, and Heng]{MM23_UDUN}
Jialun Pei, Zhangjun Zhou, Yueming Jin, He~Tang, and Pheng-Ann Heng.
\newblock Unite-divide-unite: Joint boosting trunk and structure for high-accuracy dichotomous image segmentation.
\newblock In \emph{ACM MM}, pages 2139--2147, 2023.

\bibitem[Qin et~al.(2020)Qin, Zhang, Huang, Dehghan, Zaiane, and Jagersand]{PR20_U2Net}
Xuebin Qin, Zichen Zhang, Chenyang Huang, Masood Dehghan, Osmar~R Zaiane, and Martin Jagersand.
\newblock U2-net: Going deeper with nested u-structure for salient object detection.
\newblock \emph{Pattern Recognition}, 106:\penalty0 107404, 2020.

\bibitem[Qin et~al.(2022)Qin, Dai, Hu, Fan, Shao, and Van~Gool]{DIS5K}
Xuebin Qin, Hang Dai, Xiaobin Hu, Deng-Ping Fan, Ling Shao, and Luc Van~Gool.
\newblock Highly accurate dichotomous image segmentation.
\newblock In \emph{ECCV}, pages 38--56. Springer, 2022.

\bibitem[Quan et~al.(2024)Quan, Xu, Wang, Guan, and Zheng]{TGRS24_SFANet}
Yueqian Quan, Honghui Xu, Renfang Wang, Qiu Guan, and Jianwei Zheng.
\newblock Orsi salient object detection via progressive semantic flow and uncertainty-aware refinement.
\newblock \emph{IEEE Transactions on Geoscience and Remote Sensing}, 62:\penalty0 1--13, 2024.

\bibitem[Radford et~al.(2021)Radford, Kim, Hallacy, Ramesh, Goh, Agarwal, Sastry, Askell, Mishkin, Clark, et~al.]{ICML21_CLIP}
Alec Radford, Jong~Wook Kim, Chris Hallacy, Aditya Ramesh, Gabriel Goh, Sandhini Agarwal, Girish Sastry, Amanda Askell, Pamela Mishkin, Jack Clark, et~al.
\newblock Learning transferable visual models from natural language supervision.
\newblock In \emph{ICML}, pages 8748--8763. PmLR, 2021.

\bibitem[Ravi et~al.(2025)Ravi, Gabeur, Hu, Hu, Ryali, Ma, Khedr, R{\"a}dle, Rolland, Gustafson, Mintun, Pan, Alwala, Carion, Wu, Girshick, Dollar, and Feichtenhofer]{SAM2}
Nikhila Ravi, Valentin Gabeur, Yuan-Ting Hu, Ronghang Hu, Chaitanya Ryali, Tengyu Ma, Haitham Khedr, Roman R{\"a}dle, Chloe Rolland, Laura Gustafson, Eric Mintun, Junting Pan, Kalyan~Vasudev Alwala, Nicolas Carion, Chao-Yuan Wu, Ross Girshick, Piotr Dollar, and Christoph Feichtenhofer.
\newblock {SAM} 2: Segment anything in images and videos.
\newblock In \emph{ICLR}, 2025.

\bibitem[Ronneberger et~al.(2015)Ronneberger, Fischer, and Brox]{MICCAI15_UNet}
Olaf Ronneberger, Philipp Fischer, and Thomas Brox.
\newblock U-net: Convolutional networks for biomedical image segmentation.
\newblock In \emph{MICCAI}, pages 234--241. Springer, 2015.

\bibitem[Ryali et~al.(2023)Ryali, Hu, Bolya, Wei, Fan, Huang, Aggarwal, Chowdhury, Poursaeed, Hoffman, et~al.]{ICML23_Hiera}
Chaitanya Ryali, Yuan-Ting Hu, Daniel Bolya, Chen Wei, Haoqi Fan, Po-Yao Huang, Vaibhav Aggarwal, Arkabandhu Chowdhury, Omid Poursaeed, Judy Hoffman, et~al.
\newblock Hiera: A hierarchical vision transformer without the bells-and-whistles.
\newblock In \emph{ICML}, pages 29441--29454. PMLR, 2023.

\bibitem[Skurowski et~al.(2018)Skurowski, Abdulameer, B{\l}aszczyk, Depta, Kornacki, and Kozie{\l}]{chameleon}
Przemys{\l}aw Skurowski, Hassan Abdulameer, J~B{\l}aszczyk, Tomasz Depta, Adam Kornacki, and P~Kozie{\l}.
\newblock Animal camouflage analysis: Chameleon database.
\newblock \emph{Unpublished Manuscript}, 2\penalty0 (6):\penalty0 7, 2018.

\bibitem[Sun et~al.(2021)Sun, Chen, Zhou, Zhang, and Liu]{IJCAI21_C2FNet}
Yujia Sun, Geng Chen, Tao Zhou, Yi~Zhang, and Nian Liu.
\newblock Context-aware cross-level fusion network for camouflaged object detection.
\newblock In \emph{IJCAI}, pages 1025--1031, 2021.

\bibitem[Tu et~al.(2021)Tu, Wang, Li, Fan, Zhao, and Luo]{ORSI-4199}
Zhengzheng Tu, Chao Wang, Chenglong Li, Minghao Fan, Haifeng Zhao, and Bin Luo.
\newblock Orsi salient object detection via multiscale joint region and boundary model.
\newblock \emph{IEEE Transactions on Geoscience and Remote Sensing}, 60:\penalty0 1--13, 2021.

\bibitem[Wang et~al.(2024{\natexlab{a}})Wang, Vasu, Faghri, Vemulapalli, Farajtabar, Mehta, Rastegari, Tuzel, and Pouransari]{CVPRW24_SAMCLIP}
Haoxiang Wang, Pavan Kumar~Anasosalu Vasu, Fartash Faghri, Raviteja Vemulapalli, Mehrdad Farajtabar, Sachin Mehta, Mohammad Rastegari, Oncel Tuzel, and Hadi Pouransari.
\newblock Sam-clip: Merging vision foundation models towards semantic and spatial understanding.
\newblock In \emph{CVPRW}, pages 3635--3647, 2024{\natexlab{a}}.

\bibitem[Wang et~al.(2020)Wang, Sun, Cheng, Jiang, Deng, Zhao, Liu, Mu, Tan, Wang, et~al.]{PAMI20_HRNet}
Jingdong Wang, Ke~Sun, Tianheng Cheng, Borui Jiang, Chaorui Deng, Yang Zhao, Dong Liu, Yadong Mu, Mingkui Tan, Xinggang Wang, et~al.
\newblock Deep high-resolution representation learning for visual recognition.
\newblock \emph{IEEE Transactions on Pattern Analysis and Machine Intelligence}, 43\penalty0 (10):\penalty0 3349--3364, 2020.

\bibitem[Wang et~al.(2024{\natexlab{b}})Wang, Maalouf, Xiao, Ban, Amini, Rosman, Karaman, and Rus]{drive_survey}
Tsun-Hsuan Wang, Alaa Maalouf, Wei Xiao, Yutong Ban, Alexander Amini, Guy Rosman, Sertac Karaman, and Daniela Rus.
\newblock Drive anywhere: Generalizable end-to-end autonomous driving with multi-modal foundation models.
\newblock In \emph{ICRA}, pages 6687--6694. IEEE, 2024{\natexlab{b}}.

\bibitem[Wang et~al.(2022)Wang, Xie, Li, Fan, Song, Liang, Lu, Luo, and Shao]{pvtv2}
Wenhai Wang, Enze Xie, Xiang Li, Deng-Ping Fan, Kaitao Song, Ding Liang, Tong Lu, Ping Luo, and Ling Shao.
\newblock Pvt v2: Improved baselines with pyramid vision transformer.
\newblock \emph{Computational Visual Media}, 8\penalty0 (3):\penalty0 415--424, 2022.

\bibitem[Wei et~al.(2020)Wei, Wang, and Huang]{AAAI20_F3Net}
Jun Wei, Shuhui Wang, and Qingming Huang.
\newblock F$^3$net: fusion, feedback and focus for salient object detection.
\newblock In \emph{AAAI}, pages 12321--12328, 2020.

\bibitem[Wu et~al.(2025)Wu, Xiong, Gao, Li, and Chen]{SPL25_HFSSAM2}
Zihuang Wu, Xinyu Xiong, Guangwei Gao, Hongwei Li, and Hua Chen.
\newblock Hfs-sam2: Segment anything model 2 with high-frequency feature supplementation for camouflaged object detection.
\newblock \emph{IEEE Signal Processing Letters}, 2025.

\bibitem[Xie et~al.(2024)Xie, Chen, Zhang, Wan, and Li]{LMM_Agent}
Junlin Xie, Zhihong Chen, Ruifei Zhang, Xiang Wan, and Guanbin Li.
\newblock Large multimodal agents: A survey.
\newblock \emph{arXiv preprint arXiv:2402.15116}, 2024.

\bibitem[Xin et~al.(2024)Xin, Luo, Zhou, Du, Liu, Fan, Li, and Du]{arXiv24_PEFTSurvey}
Yi~Xin, Siqi Luo, Haodi Zhou, Junlong Du, Xiaohong Liu, Yue Fan, Qing Li, and Yuntao Du.
\newblock Parameter-efficient fine-tuning for pre-trained vision models: A survey.
\newblock \emph{arXiv preprint arXiv:2402.02242}, 2024.

\bibitem[Xiong et~al.(2024)Xiong, Wu, Tan, Li, Tang, Chen, Li, Ma, and Li]{sam2unet}
Xinyu Xiong, Zihuang Wu, Shuangyi Tan, Wenxue Li, Feilong Tang, Ying Chen, Siying Li, Jie Ma, and Guanbin Li.
\newblock Sam2-unet: Segment anything 2 makes strong encoder for natural and medical image segmentation.
\newblock \emph{arXiv preprint arXiv:2408.08870}, 2024.

\bibitem[Xu et~al.(2025{\natexlab{a}})Xu, Zheng, Wang, Zhang, Ren, Xu, and Xu]{InfoFus25_SAMamba}
Wenhao Xu, Shuchen Zheng, Changwei Wang, Zherui Zhang, Chuan Ren, Rongtao Xu, and Shibiao Xu.
\newblock Samamba: Adaptive state space modeling with hierarchical vision for infrared small target detection.
\newblock \emph{Information Fusion}, page 103338, 2025{\natexlab{a}}.

\bibitem[Xu et~al.(2025{\natexlab{b}})Xu, Yang, and Xu]{arXiv25_DSUNet}
Yimin Xu, Fan Yang, and Bin Xu.
\newblock Dsu-net: An improved u-net model based on dinov2 and sam2 with multi-scale cross-model feature enhancement.
\newblock \emph{arXiv preprint arXiv:2503.21187}, 2025{\natexlab{b}}.

\bibitem[Yin et~al.(2024)Yin, Fu, Zhao, Li, Sun, Xu, and Chen]{MLLM_Survey}
Shukang Yin, Chaoyou Fu, Sirui Zhao, Ke~Li, Xing Sun, Tong Xu, and Enhong Chen.
\newblock A survey on multimodal large language models.
\newblock \emph{National Science Review}, 11\penalty0 (12):\penalty0 nwae403, 2024.

\bibitem[Zeng et~al.(2023)Zeng, Xu, Hu, Tang, Hu, and Nie]{TGRS23_AESINet}
Xiangyu Zeng, Mingzhu Xu, Yijun Hu, Haoyu Tang, Yupeng Hu, and Liqiang Nie.
\newblock Adaptive edge-aware semantic interaction network for salient object detection in optical remote sensing images.
\newblock \emph{IEEE Transactions on Geoscience and Remote Sensing}, 61:\penalty0 1--16, 2023.

\bibitem[Zhang et~al.(2020)Zhang, Cong, Li, Cheng, Fang, Cao, Zhao, and Kwong]{TIP20_EORSSD}
Qijian Zhang, Runmin Cong, Chongyi Li, Ming-Ming Cheng, Yuming Fang, Xiaochun Cao, Yao Zhao, and Sam Kwong.
\newblock Dense attention fluid network for salient object detection in optical remote sensing images.
\newblock \emph{IEEE Transactions on Image Processing}, 30:\penalty0 1305--1317, 2020.

\bibitem[Zhao et~al.(2024)Zhao, Pang, Zhang, Lu, and Zhang]{gatenet}
Xiaoqi Zhao, Youwei Pang, Lihe Zhang, Huchuan Lu, and Lei Zhang.
\newblock Towards diverse binary segmentation via a simple yet general gated network.
\newblock \emph{International Journal of Computer Vision}, 132\penalty0 (10):\penalty0 4157--4234, 2024.

\bibitem[Zheng et~al.(2024)Zheng, Gao, Fan, Liu, Laaksonen, Ouyang, and Sebe]{AIR24_BiRefNet}
Peng Zheng, Dehong Gao, Deng-Ping Fan, Li~Liu, Jorma Laaksonen, Wanli Ouyang, and Nicu Sebe.
\newblock Bilateral reference for high-resolution dichotomous image segmentation.
\newblock \emph{CAAI Artificial Intelligence Research}, 3, 2024.

\bibitem[Zhou et~al.(2022{\natexlab{a}})Zhou, Shen, Liu, Gong, Zhang, and Yan]{TGRS22_EMFINet}
Xiaofei Zhou, Kunye Shen, Zhi Liu, Chen Gong, Jiyong Zhang, and Chenggang Yan.
\newblock Edge-aware multiscale feature integration network for salient object detection in optical remote sensing images.
\newblock \emph{IEEE Transactions on Geoscience and Remote Sensing}, 60:\penalty0 1--15, 2022{\natexlab{a}}.

\bibitem[Zhou et~al.(2022{\natexlab{b}})Zhou, Shen, Weng, Cong, Zheng, Zhang, and Yan]{TCYB22_ERPNet}
Xiaofei Zhou, Kunye Shen, Li~Weng, Runmin Cong, Bolun Zheng, Jiyong Zhang, and Chenggang Yan.
\newblock Edge-guided recurrent positioning network for salient object detection in optical remote sensing images.
\newblock \emph{IEEE Transactions on Cybernetics}, 53\penalty0 (1):\penalty0 539--552, 2022{\natexlab{b}}.

\bibitem[Zhu et~al.(2025)Zhu, Wang, Chen, Liu, Ye, Gu, Tian, Duan, Su, Shao, et~al.]{internvl3}
Jinguo Zhu, Weiyun Wang, Zhe Chen, Zhaoyang Liu, Shenglong Ye, Lixin Gu, Hao Tian, Yuchen Duan, Weijie Su, Jie Shao, et~al.
\newblock Internvl3: Exploring advanced training and test-time recipes for open-source multimodal models.
\newblock \emph{arXiv preprint arXiv:2504.10479}, 2025.

\end{thebibliography}
\end{document}